\documentclass{aamas2017}

\usepackage{algorithmic}
\usepackage{algorithm}
\usepackage{array}
\usepackage{eqparbox}

\newcommand{\RN}[1]{%
  \textup{\uppercase\expandafter{\romannumeral#1}}%
}

\begin{document}


\title{Traffic Light Control Using Deep Policy-Gradient and Value-Function Based Reinforcement Learning}

\numberofauthors{4}

\author{
%
\alignauthor
Seyed Sajad Mousavi\\
    \affaddr{Discipline of IT}\\
       \affaddr{National University of Ireland, Galway}\\
       \email{s.mousavi1@nuiglaway.ie}
\alignauthor
Michael Schukat\\
       \affaddr{Discipline of IT}\\
       \affaddr{National University of Ireland, Galway}\\
       \email{michael.schukat@nuigalway.ie}
\and 
\alignauthor
Enda Howley\\
       \affaddr{Discipline of IT}\\
       \affaddr{National University of Ireland, Galway}\\
       \email{enda.howley@nuigalway.ie}       
}

\maketitle

\begin{abstract}

Recent advances in combining deep neural network architectures with reinforcement learning techniques have shown promising potential results in solving complex control problems with high dimensional state and action spaces. Inspired by these successes, in this paper, we build two kinds of reinforcement learning algorithms: deep policy-gradient and value-function based agents which can predict the best possible traffic signal for a traffic intersection. At each time step, these adaptive traffic light control agents receive a snapshot of the current state of a graphical traffic simulator and produce control signals. The policy-gradient based agent maps its observation directly to the control signal, however the value-function based agent first estimates values for all legal control signals. The agent then selects the optimal control action with the highest value. Our methods show promising results in a traffic network simulated in the SUMO traffic simulator, without suffering from instability issues during the training process.

\end{abstract}






\section{Introduction}
\label{introduction}

With regard to fast growing population around the world, the urban population in the 21\textsuperscript{st} century is expected to increase dramatically. Hence, it is imperative that urban infrastructure is managed effectively to contend with this growth. One of the most critical consideration when designing modern cities is developing smart traffic management systems. The main goal of a traffic management system is reducing traffic congestion which nowadays is one of the major issues of megacities. Efficient urban traffic management results in timea and financial savings as well as reducing CO\textsubscript{2} emission into atmosphere. To address this issue, a lot of solutions have been proposed \cite{li2014survey,balaji2010urban,abdoos2013holonic,li2016parallel}. They can be roughly classified into three groups. The first is pre-timed signal control, where a fixed time is determined for all green phases according to historical traffic demand, without considering possible fluctuations in traffic demand. The second is vehicle-actuated signal control where, traffic demand information is used, provided by inductive loop detectors on an equipped intersection to decide to control the signals, e.g. extending or terminating a green phase. The third is adaptive signal control, where the signal timing control is managed and updated automatically according to the current state of the intersection (i.e. traffic demand, queue length of vehicles in each lane of the intersection and traffic flow fluctuation) \cite{el2013multiagent}. In this study, we are interested in the third approach and aim to propose two novel methods for traffic signal control by leveraging recent advances in machine learning and artificial intelligence fields \cite{mnih2013playing,mnih2015human}.

Reinforcement learning \cite{Sutton98} as a machine learning technique for traffic signal control problem has led to impressive results \cite{balaji2010urban,prashanth2011reinforcement} and has shown a promising potential solver. It does not need to have a perfect knowledge of the environment in advance, for example traffic flow. Instead they are able to gain knowledge and model the dynamics of the environment just by interacting with it. A reinforcement learning agent learns based on trial and error. It receives a scalar reward 
after taking each action in the environment. The obtained reward is based on how well the taken action is and the agent's goal is to learn an optimal control policy so the discounted cumulative reward is maximized via repeated interaction with its environment. Aside from traffic control, reinforcement learning has been applied to a number of real world problems such as cloud computing \cite{duggan2016reinforcement,duggan2016autonomous}.

Typically the complexity of using reinforcement learning in real world applications such as traffic signal management, grows exponentially as state and action spaces increase. To deal with this problem, function approximation techniques and hierarchical reinforcement learning approaches can be used.  Recently, deep learning has gained huge attraction and has been successfully combined with reinforcement learning techniques to deal with complex optimization problems such as playing Atari 2600 games \cite{mnih2015human}, Computer Go program \cite{silver2016mastering}, etc., where the classical RL methods could not provide optimal solutions. In this way, the current state of the environment is fed into a deep neural net (e.g. a convolutional neural network \cite{lecun1998gradient}) trained by reinforcement learning techniques to predict the next possible optimal action(s). 

Inspired by the successes of combining reinforcement learning with deep learning paradigm and with regard to the complex nature of environment of traffic signal control problem, in this paper we aim to use the effectiveness and power of deep reinforcement learning to build  adaptive signal control methods in order to optimize the traffic flow. Although a few previous studies have tried to apply deep reinforcement learning in the traffic signal control problem \cite{vancoordinated2016,genders2016using}, in this research the state representation is different. Also, One of our methods uses policy gradient method which does not suffer from oscillations and instabilities during training process and can take full advantage of the available data of the environment to develop the optimal control policy.

We propose adaptive signal controllers by combination two reinforcement learning approaches (i.e. policy gradient and action-value function) and a deep convolution neural network, which perceive embedded camera observations in order to produce control signals in an isolated intersection. We conduct simulated experiments with our proposed methods in SUMO traffic simulator. 

The rest of the paper is organized as follows. Section \ref{related-work}  provides related work in the area of traffic light control (TLC). Section \ref{background} gives a brief review of reinforcement learning techniques which we have used in this research. Section \ref{model} presents how to formulate the TLC problem as a reinforcement learning task and the proposed methods to solve the task. Then Section \ref{results} provides simulation results and the performance of the proposed approaches. Finally Section \ref{conclusion} concludes the paper and give some directions for future research.

\section{Related work}
\label{related-work}

A lot of research has been done in academic and industry communities to build adaptive traffic signal control systems. In particular, significant research has been conducted employing reinforcement learning methods in the area of traffic light signal control \cite{wiering2000multi,abdulhai2003reinforcement,brockfeld2001optimizing}. These works have achieved promising results. However, their simulation testbeds have not been mature enough to be comparable with more realistic situations. Developing advance traffic simulation tools have made researchers develop novel state representation and reward functions for reinforcement learning algorithms, which could consider more aspects of complexity and reality of real-world traffic problems \cite{el2013multiagent,abdoos2013holonic,chin2011q,arel2010reinforcement}. All this these attempts viewed the traffic light control problem as a fully observable Markov decision process (MDP) and investigated whether Q-learning algorithm can be applied to it. However, Richter's study formulated the traffic problem as a partially observable MDP (POMDP) and applied policy gradient methods to guarantee local convergence under a partial observable environment \cite{ritcher2007traffic}.

By utilizing advances in deep learning and its application to different domains \cite{deng2014tutorial,duggan2016autonomous}, deep learning has gained attention in the area of traffic management systems. Previous research has used deep stacked autoencoders (SAE) neural networks to estimate Q-values, where each Q-value is corresponding to each available signal phase \cite{li2016traffic}. It considered measures of speed and queueing length as its state in each time step of learning process of its proposed method. Two recent studies by \cite{vancoordinated2016,genders2016using} provided deep reinforcement learning agents that used deep Q-netwok \cite{mnih2015human} to map from given states to Q-values. Their state representations were a binary matrix of the positions of vehicles on the lanes of an intersection, and a combination of the presence matrix of vehicles, speed and the current traffic signal phase, respectively. However, we use raw visual input data of the traffic simulator snapshots as system states. Moreover, in addition to estimating Q-function, one of the proposed methods directly maps from the input state to a probability distribution over actions (i.e. signal phases) via deep policy gradient method.

\section{Background}
\label{background}
In this section, we will review Reinforcement Learning (RL) approaches and briefly describe how RL is applied to real world problems where the number of states and actions are extremely high so that the regular reinforcement learning techniques cannot deal with them.

\subsection{Reinforcement Learning}

\begin{figure}
\centering
\includegraphics[width=\linewidth]{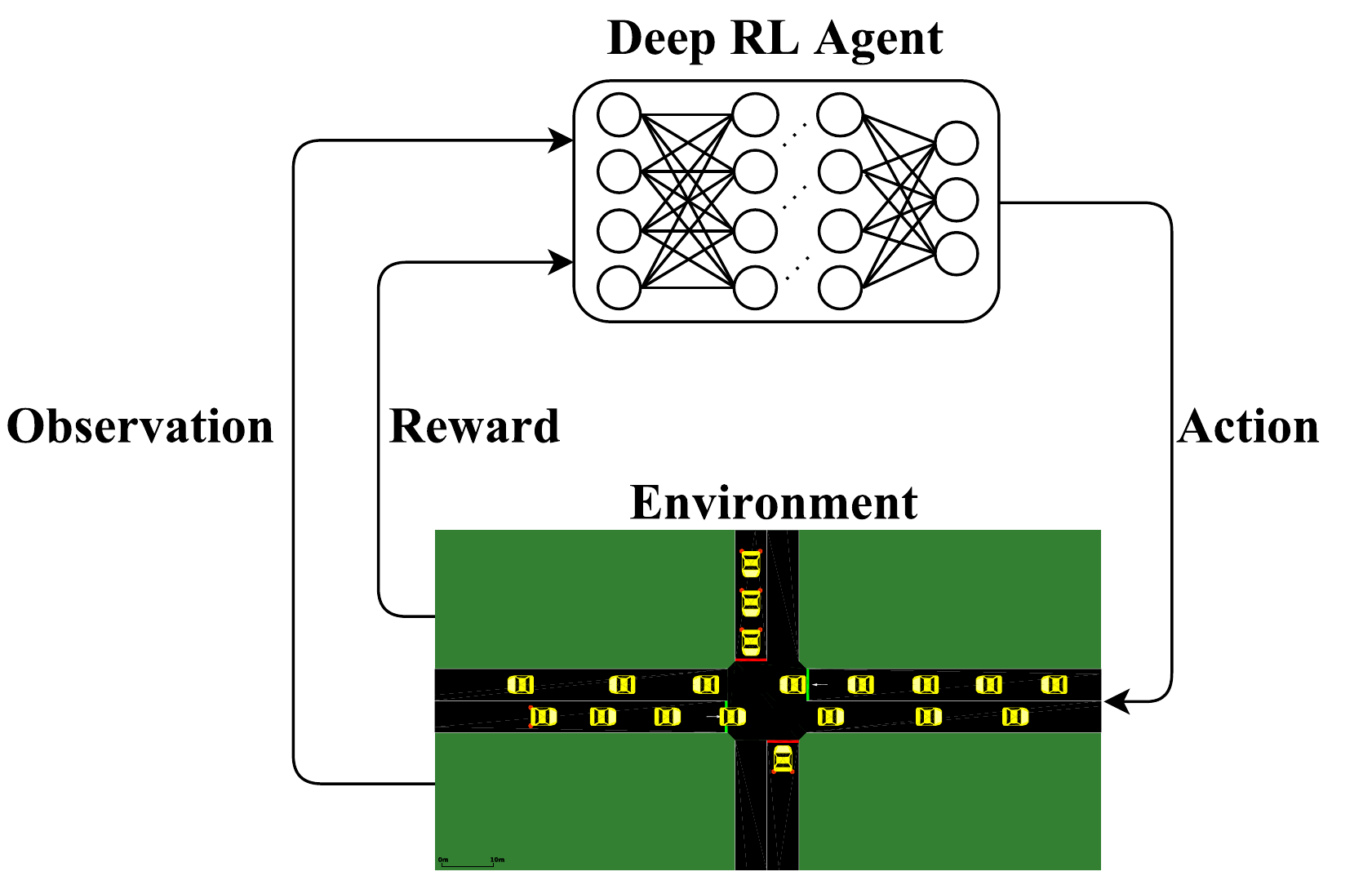}
\caption{Deep reinforcement learning agent of traffic signal control.}
\label{fig:f1}
\end{figure}

A common reinforcement learning \cite{Sutton98} setting is shown in Figure \ref{fig:f1} where an RL agent interacts with an environment. The interaction is continued until reaching a terminal state or the agent meets a termination condition. Usually the problems that RL techniques are applied to, are treated as Markov decision processes (MDPs). A MDP is defined as a five-tuple $<S,A,T,R,\gamma>$ , where S is the set of states in the state space of the environment, A  is the set of actions in the action space that the agent can use in order to interact with the environment, T is the transition function, which is the probability of moving between the environment states, R is the reward function and $\gamma \in [0,1]$ is known as the discount factor, which models the importance of the future and immediate rewards. At each time step $t$, the agent perceives the state $s_t \in S$ and, based on its observation, selects an action $a_t$. Taking the action, leads to the state of the environment transitions to the next states $ s_{t+1} \in S$ regarding the transition function $T$. Then, the agent receives reward $r_t$ which is determined by the reward function $R$.

The goal of the learning agent in reinforcement learning framework is to learn an optimal policy $ \pi: S \times A \to [0, 1]$ which defines the probability of selecting action $a_t$ in state $s_t$, so that with following the underlying policy the expected cumulative discounted reward over time is maximized. The discounted future reward, $R_t$ at time $t$ is defined as follows:
\begin{equation}R_t=E[\sum_{k=0}^{\infty} \gamma^k r_{t+k}], \end{equation}
where the role of the discount factor $\gamma$ is to trade off the worth of immediate and future rewards.
In most real world problems, there are many states and actions which make it impossible to apply classic reinforcement learning techniques, which consider tabular representations for their state and action spaces. For example, in the problem of traffic light optimization, that we interest in this paper, the state space is continuous. 
Hence, it is common to use function approximators \cite{xu2014} or decomposition and aggregation techniques like Hierarchical Reinforcement Learning (HRL) approaches \cite{barto2003recent,ghazanfari2014enhancing,mousavi2014automatic} and advance HRL \cite{ghazanfari2016extracting}. 

Different forms of function approximators can be used with reinforcement learning techniques. For example, linear function approximation, a linear combination of feature of state and action spaces $f$ and learned weights $w$ (e.g. $\sum_{i}f_iw$) or a non-linear function approximation (e.g. a neural network). Until recently, the majority of work in reinforcement learning has been applying linear function approximatiors. More recently, deep neural networks (DNNs) such as convolutional neural networks (CNNs), recurrent neural networks (RNNs), stacked auto-encoders (SAE), etc. have also been commonly used as function approximators for large reinforcement learning tasks \cite{lange2010deep,mnih2013playing}. The interested readers are referred to \cite{mousavi2016deep} for a review of using deep neural networks with reinforcement learning framework.

\subsection{Deep learning and Deep Q-learning}

Deep learning techniques are one of the best solutions to address high dimensional data and extract discriminative information from the data. Deep learning algorithms have the capability of automating feature extraction (the extraction of representations) from the data. The representation are learnt through the data which are fed directly into deep nets without using human knowledge (i.e. automated feature extraction). Deep learning models contain multiple layers of representations. Indeed, it is a stack of building blocks such as auto-encoders, Restricted Boltzmann Machines (RBMs) and convolutional layers. During training, the raw data is fed into a network consisting of multiple layers. The output of the each layer which is nonlinear feature transformations, is used as inputs to the next layers of the deep neural network. The output representation of the final layer can be used for constricting classifiers or those applications which can have the better efficiency and performance with abstract representation of the data in a hierarchical manner as inputs. A nonlinear transformation is applied at each layer on its input to try to learn and extract underlying explanatory factors. Consequently, this process learns a hierarchy of abstract representations.

One of the main advantages of deep neural networks is the capability of automating feature extraction from row input data. A deep Q-learning Network (DQN) \cite{mnih2013playing} uses this benefit of deep learning in order to represent the agent's observation as an abstract representation in learning an optimal control policy. The DQN method aggregates a deep neural network function approximator with Q-learning to learn action value function and as a result a policy $\pi$, the behaviour of the agent which tells the agent what action should be selected for each input state. Applying non-linear function approximators such as neural networks with model-free reinforcement learning algorithms in high-dimensional continuous state and action spaces, has some convergence problems \cite{tsitsiklis1997analysis}. The reasons for these issues are: 1) Consecutive states in reinforcement learning tasks have correlation. 2) The underlying policy of the agent is changing frequently, because of slight changes in Q-values. To cope with these problems, the DQN provides some solutions which improve the performance of the algorithm significantly. For the problem of correlated states, DQN uses the previously proposed experience replay approach \cite{lin1993reinforcement}. In this way, at each time step, the DQN stores the agent's experience $(s_t,a_t,r_t,r_{t+1})$ into a date set $D$, where $s_t$, $a_t$, and $r_t$ are the state, chosen action and received reward, respectively and $s_{t+1}$ is the state at the next time step. To update the network, the DQN utilizes stochastic minibatch updates with uniformly random sampling from the experience replay memory (previous observed transitions) at training time. This negates strong correlations between consecutive samples. Another approach to deal with aforementioned convergence issues, which we also examine in this research, is the policy gradient methods. This approach has demonstrated better convergence properties in some RL problems \cite{sutton1999policy}.

\subsection{Policy Gradient Methods}
A Policy Gradient (PG) method tries to optimize a parameterized 
policy function by gradient descent method. Indeed, policy gradient methods are interested in searching policy space to learn policies directly, instead of estimating state-value or action-value functions. Unlike the traditional reinforcement learning algorithms, PG methods do not suffer from the convergence problems of estimating value functions under nonlinear function approximation or in the environments which might be partially observable MDPs. They can also deal with the complexity of continuous state and action spaces better than purely value-based methods \cite{sutton1999policy}.  Policy gradient methods estimate policy gradients using Monte Carlo estimates of the policy gradients \cite{baxter2001experiments}.  These methods are guaranteed to converge to a local optimum of their parametrized policy function. However, typically PG methods result in high variance in their gradient estimates. Hence, in order to reduce the variance of the gradient estimators, some methods subtract a base line function from the policy gradients. The baseline function can be calculated in different manners \cite{schulman2015gradient,wierstra2010recurrent}. By inspiring these features of PG methods and successes of neural networks in automatic feature abstractions, we use deep neural networks to represent an optimal traffic control policy directly in the traffic signal control problem.

\section{System Description}
\label{model}
In this section, we will formulate traffic light control problem as a reinforcement learning task by describing the states, actions and reward function. We then present the policy as a deep neural network and how to train the network.

\subsection{State Representation}
We represent the state of the system as an image $s_t \in R^d$ or a snapshot of the current state of a graphical simulator (e.g. SUMO-GUI \cite{krajzewicz2012recent}) which is a vector of row pixel values of current view of the intersection at each step of simulation (as shown in Figure \ref{fig:f1}). This kind of representation is like putting a camera on an intersection which enables it to view the whole intersection. The state representation in the traffic light control literature usually uses a vector representing the presence of a vehicle at the intersection, a Boolean-valued vector where a value 1 indicates the presence of a vehicle and a value 0 indicates the absence of a vehicle \cite{vancoordinated2016,thorpe1996tra}, or a combination of the presence vector with another vector indicating the vehicle's speed at the given intersection \cite{genders2016using}. Regardless of these states representations that are using a prior knowledge provided, they make assumptions which are not generalizable for the real world. For instance, they discretize a lane segment of an intersection into cells with a constant length $c$ which is supposed to be the vehicle length to build the vehicle's speed and presence vectors. However, by feeding the state as an image to a convolutional neural network, the system can detect the location and presence of all vehicles with different lengths and as result the vehicles' queue on each lane. Furturmore, by stacking a history of consecutive observations as input, the convolutional layers of a deep network are able to estimate velocity and travel direction of vehicles. Hence, the system can implicitly benefit from these information as well. 

\subsection{Action Set}
To control traffic signal phases, we define a set of possible actions $A$ = \{North/South Green (NSG), East/West Green (EWG)\}. NSG allows vehicles to pass from North to South and vice versa, and also indicates the vehicles on East/West route should stop and not proceed through the intersection. EWG allows vehicles to pass from East to West and vice versa, and implies the vehicles on North/South route should stop and not proceed through the intersection. At each time step t, an agent regarding its strategy chooses an action $a_t \in A$. Depending the selected action, the vehicles on each lane are allowed to cross the intersection.

\subsection{Reward Function}
Typically an immediate reward $r_t \in 
\mathbb{R}$ is a scalar value which the agent receives after taking the chosen action in the environment at each time step. We set the reward as the difference between the total cumulative delays of two consecutive actions, i.e.

\begin{equation} r_t = D_{t-1} - D_t, \end{equation}
where $D_t$ and $D_{t-1}$ are the total cumulative delays in the current and previous time steps. The total cumulative delay at time t, is the summation of the cumulative delay of all the vehicles appeared from $t = 0$ to current time step t in the system. The positive reward values imply the taken actions led to decrease the total cumulative delay and the negative rewards imply an increase in the delay. With regard to the reward values, the agent may decide to change its policy in certain states of the system in the future.

\subsection{Agent's Policy}
The agent chooses the actions based on a policy $\pi$. In the policy-based algorithm, the policy is defined as a mapping from the input state to a probability distribution over actions $A$. We use the deep neural network as the function approximator and refer its parameters $\theta$ as policy parameters. The policy distribution $\pi(a_t |s_t;\theta)$ is learned by performing gradient descent on the policy parameters. In the value-function based algorithm, the deep neural network is utilized to estimate the action-value function. The action-value function maps the input state to action values, which each represents the future reward that can be achieved for the given state and action. The optimal policy can then be extracted by performing a greedy approach to select the best possible action. 

\begin{algorithm} \label{alg1}
\caption{Deep Value-Function based reinforcement learning agent of traffic signal control with experience replay}
\begin{algorithmic}[1]
\STATE Initialize parameters, $\theta$ with random values
\STATE Initialize replay memory $M$ with capacity $L$
\FOR {each simulation}
 \STATE initialize $s$ with current view of the intersection
 \REPEAT[each step in the simulation]
   \STATE choose action $a$ according to $\epsilon$-greedy policy
   \STATE take action $a$, observe reward $r$ and next state $s'$
   \STATE store transition $(s,a,r,s')$ in $M$
   \STATE $s\gets s'$
   \STATE $b\gets$ sample random minbatch of transitions from the replay memory, $M$
   \FOR {each transition $(s_j,a_j,r_j,s_j')$ in $b$}
     \IF{$s_j'$ is terminal}
       \STATE  $y_i \gets r_j$
     \ELSE
     \STATE  $y_j  = r_j + \gamma max_{a'} Q({s'}_j,a';{{\theta}^-}_{i-1})$ 
     \ENDIF
     \STATE update parameters $\theta$ according to equation \eqref{eq:eq7} 
    \ENDFOR
 \UNTIL {$s$ is terminal}
\ENDFOR
 \end{algorithmic}
\end{algorithm}

\subsection{Objective Function and System Training}
There are many measures such as maximizing throughput, minimizing and balancing queue length, minimizing the delay, etc. in the traffic signal management literature to consider as the learning agent's objective function. In this research, the agent aims to maximize the reduction in the total cumulative delay, which empirically has been shown to maximize throughput and to reduce queue length (more details discussed in Section \ref{sec:results}).

The objective of agent is to maximize the expected cumulative discounted reward. We aim to maximize the reward under the probability distribution $\pi(a_t |s_t;\theta)$:
\begin{equation} \label{eq:eq3} J(\theta) =E_{\pi_\theta} [\sum_{t=0}^{T}\gamma^t r_t] = E_{\pi_\theta}[R]. \end{equation}

We divide the system training based two RL approaches: Value-function based and Policy-based. in \textbf{value-function based approach}, the value function, $Q_{\pi}(s,a)$ is defined as follows:
\begin{equation} Q_{\pi}(s,a) = E_{\pi}[r_t + \gamma max_{a'} Q(s',a')|s,a] \end{equation}

Where it is implicit that $s, s'\in S $ and $a\in A$. the value function can be parameterized, $Q(s,s;\theta)$ with parameter vector $\theta$. Typically, the gradient-descent methods are used to learn parameters, $\theta$ by trying to minimize the following loss function of mean-squared error in Q values:
\begin{equation} \label{eq:eq5} J(\theta) = E_{\pi}[(r + \gamma max_{a'} Q(s',a';\theta)- Q(s,a;\theta))^2] \end{equation}

Where $r + \gamma max_{a'} Q(s',a';\theta)$ is the target value. In the DQN algorithm, a target Q-network is used to address the instability problem of the policy. The network is trained with the target Q-network to obtain consistent Q-learning targets by keeping the weight parameters (${\theta}^-$) used in the Q-learning target fixed and updating them periodically every N steps through the parameters of the main network, $\theta$. The target value of the DQN is represented as follows:

\begin{equation}  y_i  = r + \gamma max_{a'} Q(s',a';{{\theta}^-}_{i-1}) \end{equation}

Where ${\theta}^-$ is parameters of the target network. The stochastic gradient descent method is used in order to optimize equation \eqref{eq:eq5}. The parameters of the deep Q-learning algorithm are updated as follows:

\begin{equation} \label{eq:eq7}  {\theta}_i  \leftarrow {\theta}_{i-1} + \alpha(y_i - Q(s,a;{\theta}_i)) \bigtriangledown_{{\theta}_i} Q(s,a;{\theta}_i) \end{equation}

Where $y_i$ is the target value for iteration $i$ and $\alpha$ is a scalar learning rate. Algorithm~\ref{alg1} presents the pseudo-code for the training algorithm.

In \textbf{policy-based approach}, The gradient of the objective function represented in equation \eqref{eq:eq3} is given by:

\begin{equation} \label{eq:eq8}
\bigtriangledown_{\theta} J = \sum_{t=0}^{T} E_{\pi_\theta} [\bigtriangledown_\theta log(a_t |s_t;\theta) R_t]. \end{equation}

\begin{figure}
\centering
\includegraphics[width=0.8\linewidth, height=2in]{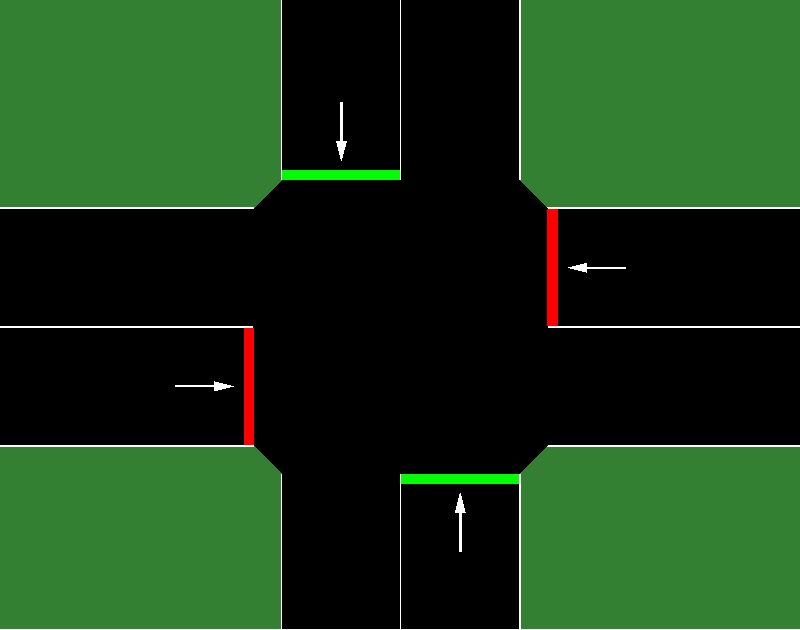}
\caption{The intersection geometry for the traffic simulation.}
\label{fig:int-geo}
\end{figure} 

This equation \eqref{eq:eq8} is standard learning rule of the REINFORCE algorithm \cite{williams1992simple}. It updates the policy parameters $\theta$  in the direction $\bigtriangledown_\theta log(a_t |s_t;\theta)$ so that the probability of action $a_t$ at state $s_t$ is increased if it has led to high cumulative reward, however it is decreased if the action has result in a low reward. The gradient estimate in equation 2 results to have high variance. It is common to reduce the variance by subtracting a baseline function $b_t(s_t)$ from the return $R_t$, without changing expectation. Commonly an estimate of the state value function is used as the baseline, $b_t(s_t ) = V^{\pi_{\theta_v}}(s_t )$. Thus, the adjusted gradient is $\bigtriangledown_\theta log(a_t |s_t;\theta)(R_t - b_t(s_t))$. The value $R_t - b_t$ is known as the \textit{advantage function}.

With regard to the advantage actor-critic method \cite{mnih2016asynchronous}, computing a single update is done by selecting actions using the underlying policy for up to $M$ steps or till a terminal state is met. In this way, the agent obtains up to $M$ rewards from the environment at each update point and updates the policy parameters after every $n \leq M $ steps regarding n-step returns. The vector parameters $\theta$ is updated through the stochastic gradient descent method:

\begin{equation} \theta \leftarrow \theta + \alpha \sum_{t}\bigtriangledown_\theta log(a_t |s_t;\theta)A(s_t,a_t;\theta,\theta_v), \end{equation}

where $A(s_t,a_t;\theta,\theta_v )$ is an estimate of the \textit{advantage function} corresponding $\sum_{i=0}^{n-1} \gamma^i r_{t+i} + \gamma^n V(s_{t+n};\theta) - V(s_t;\theta_v)$,  where $n$ might have different values with respect to the state, up to $M$. this process is an actor-critic algorithm,  the policy $\pi(a_t |s_t;\theta)$ refers to the actor and the estimate of the state value function $V^{\pi_{\theta_v}}(s_t)$ implies to the critic \cite{degris2012model,mnih2016asynchronous}. Algorithm~\ref{alg2} shows the
pseudo-code for the training algorithm.

\begin{algorithm} \label{alg2}
\caption{Deep Policy-Gradient based reinforcement learning agent of traffic signal control}
\begin{algorithmic}[1]
\STATE Initialize parameters, $\theta$, $\theta_v$ with random values
\STATE Initialize step counter $t \gets 0$
\FOR {each simulation}
 \STATE initialize $s$ with current view of the intersection
 \STATE $t_{start} = t$
 \REPEAT
   \STATE perform action $a$ according to policy $\pi(a|s;\theta)$
   \STATE observe reward $r$ and next state $s'$
   \STATE $t \gets t + 1$
 \UNTIL {$s$ is terminal \textbf{or} $t - t_{start} == M$ (Max step)}  
 \IF{$s$ is terminal}
   \STATE $R = 0$
   \ELSE
   \STATE $R = V(s;\theta_v)$
 \ENDIF
 \FOR [$n <= M$ times] {$i \in \{t-1, ..., t_{start}\}$}
    \STATE $R \gets r_i + \gamma R$
    \STATE $\theta \gets \theta + \alpha \bigtriangledown_\theta log(a_i |s_i;\theta)(R- V(s_i;\theta_v))$
    \STATE $\theta_v \gets \theta_v + \frac{\partial(R- V(s_i;\theta_v))^2}{\partial \theta_v}$
  \ENDFOR
\ENDFOR
 \end{algorithmic}
\end{algorithm}

\section{Experiment and Results}
\label{results}
In this section, we present the simulation environment, where our experiments have been done. We then describe the details of the deep neural network utilised, including hyper-parameters to represent the agent's policy.

\subsection{Experiment Setup}
We have used the Simulation of Urban MObility (SUMO) \cite{krajzewicz2012recent} tool to simulate traffic in all experiments. SUMO is a well-known open source traffic simulator which provides useful Application Programming Interfaces (APIs) and a Graphical User Interface (GUI) view to model large road networks as well as some possibilities to handle them. In particular, we utilised SUMO-GUI v0.28.0. as it allows to have snapshots of each step of the simulation. The intersection geometry used in this study is shown in Figure \ref{fig:int-geo}. There are 4 incoming lanes to the intersection and four outgoing lanes from the intersection. To generate traffic demands from different directions (i.e. north-to-south and west-to-east and vice versa) to the road network, a uniform probability distribution with the probability 0.1 was used.
\begin{table*}[htbp]
\centering
\caption{Comparison of performance of the proposed methods against the SNN method}
\label{tab:tab1}
\begin{tabular}{llll}
\hline
Evaluation Metric ($\mu$, $\sigma$; $n = 100$) & Policy-based & Value-function based & SNN     \\ \hline
Queue (Vehicles)                  & (1.79, 0.073)      & (1.74, 0.10)              & (5.55, 0.73) \\
Cumulative Delay (s)              & (11.25, 0.39)      & (11.01, 0.69)              & (41.40, 7.31) \\
Reward (r)                        & (6.14, 0.29)      & (7.14, 0.44)              & (1.73, 0.62) \\ \hline
\end{tabular}
\end{table*}

\begin{figure}
\centering
\includegraphics[width=\linewidth]{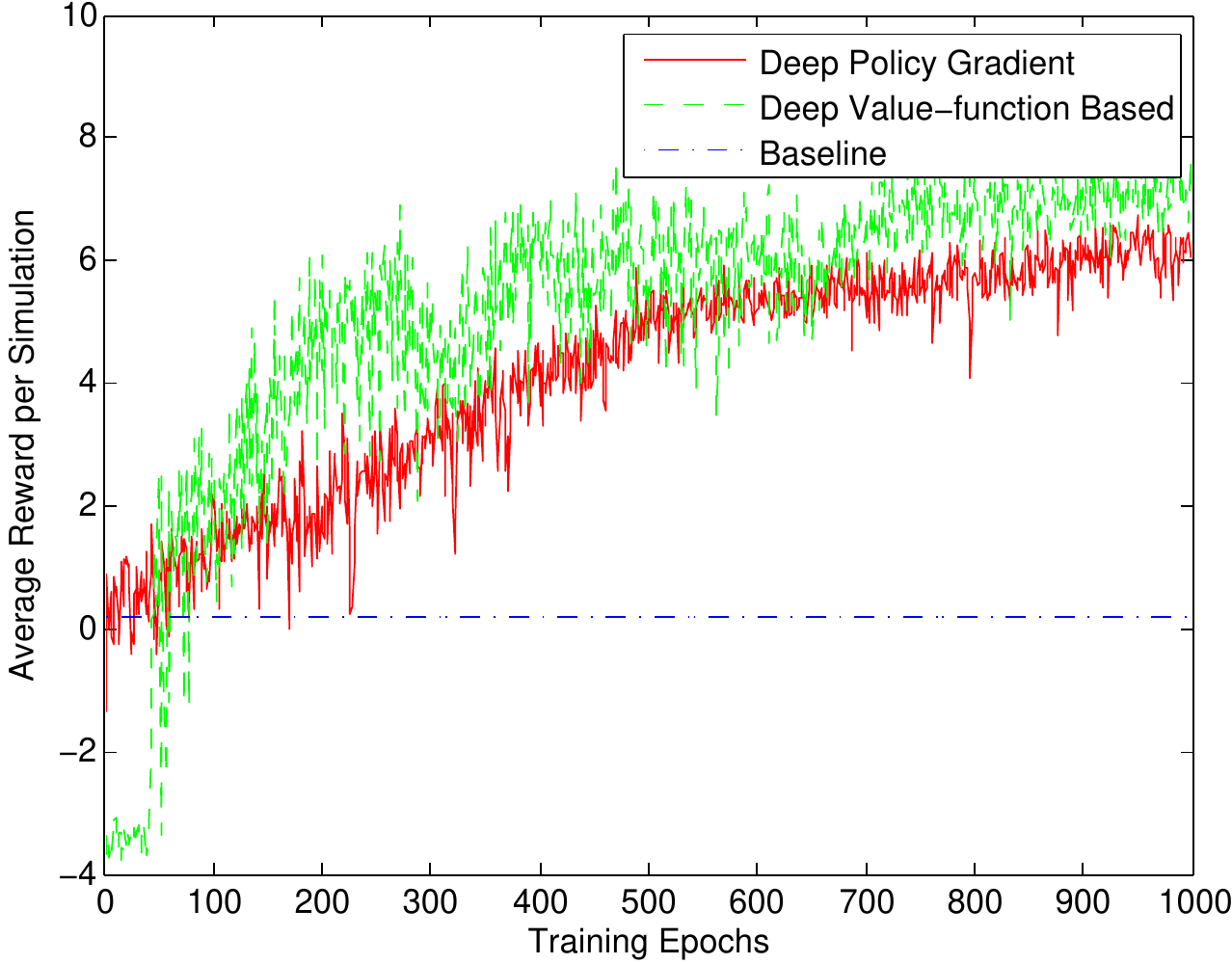}
\caption{A comparison of performance of the average reward received during the evaluation time for the proposed method and the baseline.}
\label{fig:reward}
\end{figure}
\subsection{System Architecture and Hyperparameters} \label{sec:hyper-para}
We took the snapshots from the SUMO-GUI and did some basic pre-processing. The snapshots are converted from red-green-blue (RGB) representation to gray-scale and resized them to $128 \times 128$ frames. To enable our system to memorize a history of the past observations, we stacked the last four frames of the history and provided them to the system as input. So, the input to the network was a $128\times 128 \times 4$ image.
We applied approximately the same architecture of the Deep Q-Network (DQN) algorithm introduced by Mnih et al. \cite{mnih2013playing,mnih2015human}. The network consists a stack of two convolutional layers with filters $16\ 8 \times 8$ and $32\ 4 \times 4$ and strides 4 and 2, respectively. The final hidden layer is fully-connected with 256 hidden node. All three hidden layers are followed by a rectifier nonlinearity. The main difference with the network architecture of the DQN method is the last layer, where the last layer of DQN is a fully-connected linear layer with a number of output neurons (i.e. Q-values $Q(a,s)$) corresponding to each action in a given Atari 2600 game, while in policy-based model the last layer represents two set of outputs, a softmax output resulting in a probability distribution over the actions A (i.e. the policy $ \pi(a,s)$), and a single linear output node resulting in the estimate of the state value function $V(s)$. For value-function model we used the architecture, the same as the DQN. The output layer is corresponding to action values. In all of our experiments, the discount factor was set to $\gamma = 0.99$ and all weights of the network were updated by the Adam optimizer \cite{kingma2014adam} with a learning rate $\alpha = 0.00001$ and with mini batches of size $M$ (up to 32), the maximum number of steps that the agent can take to follow its policy and afterwards needs to update it. The network was trained for about 1050 epoch, approximately 2 million time steps. Each epoch is corresponded 10 episodes and each episode was a complete SUMO-GUI simulation. The learned policies by the agent was evaluated every 10 episodes by running SUMO-GUI for 5 episodes and averaging the resulting rewards, total cumulative delay and queue length.

To evaluate our proposed method we also built a shallow neural network (SNN) with one hidden layer. The hidden layer has 64 hidden nodes followed by a rectifier nonlinearity. The output layer is a fully-connected linear layer with a number of output neurons corresponding to each traffic signal phase in the intersection. Two vectors are used as input state of the network. The first representing the number of queued vehicles at the lanes of the intersection (i.e. North, South, East and West) and the second corresponding to the current traffic signal phase of the intersection. SNN is trained with the same hyper-parameters and optimization method (i.e. the gradient decent algorithm) as the proposed methods.

\begin{figure}
\centering
\includegraphics[width=\linewidth]{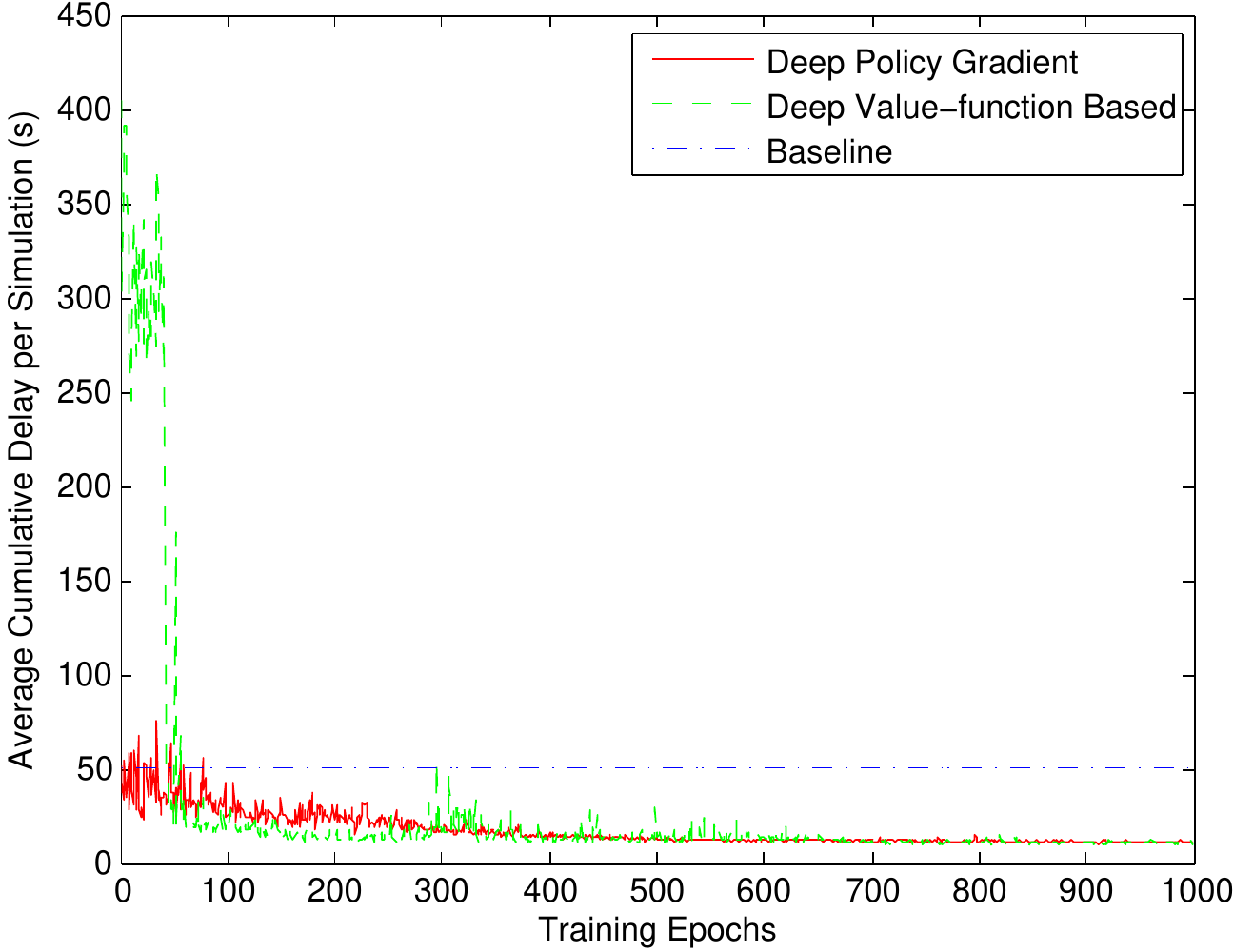}
\caption{Average Cumulative delay per vehicle using the suggested model and the baseline during the evaluation time.}
\label{fig:delay}
\end{figure} 
\begin{figure}
\centering
\includegraphics[width=\linewidth]{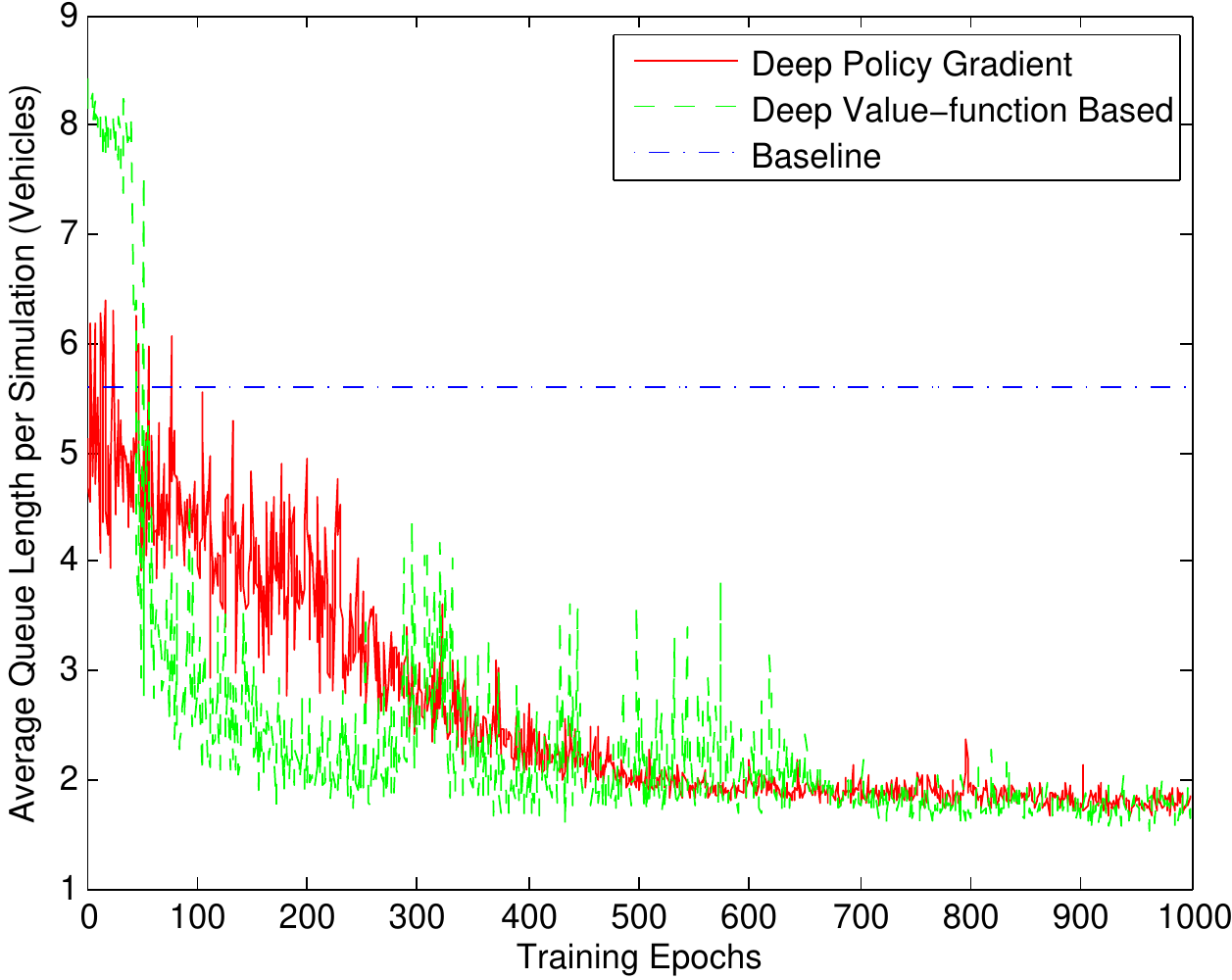}
\caption{Average queue length of the intersection using the proposed model and the baseline during the evaluation time.}
\label{fig:queue}
\end{figure} 
\subsection{Results and Discussion} \label{sec:results}
To evaluate the performance of the proposed methods, we compared them against a baseline traffic controller, a controller that gives an equal fixed time to each phase of the intersection. We ran SUMO-GUI simulator for the proposed model using the configuration setting explained in Section \ref{sec:hyper-para} and compared the average reward, average total cumulative delay and average queue length achieved to the baseline. Figure \ref{fig:reward} shows the received average reward while the agent
follows a certain policy. As shown in Figure \ref{fig:reward}, the proposed method performs significantly better than the baseline and results more reward magnitudes by doing more epochs. This gradually increasing reward reflects the agent's ability to learn an optimal control policy in a stable manner. Unlike using deep reinforcement learning for estimating the Q-values in traffic light optimisation problem \cite{vancoordinated2016}, the proposed agent doesn't suffer stability issues. In order to assess the learned policy by the agent, two of the most common performance metrics in the traffic signal control literature is implemented: the cumulative delay and queue length. Figures \ref{fig:delay} and \ref{fig:queue} illustrate the performance comparison of the leaning agent regarding average cumulative delay time and average queue length metrics, respectively, to the baseline, while the agent is following the learning policy over time. The plots clearly show the agent is able to find a policy resulting minimizing queue length and total cumulative delay. Moreover, these graphs reveal that by using the reward function for reducing cumulative delay, the intersection queue length is reduced as well as the total cumulative delay of all vehicles.

We also compared the proposed methods with the SNN, which is a shallow neural network with one hidden layer. Table \ref{tab:tab1} reports a comparison of the proposed models and the SNN model in terms of the average and standard deviation ($\mu$, $\sigma$) of  average queue length, average cumulative delay time and the received average reward metrics. The results on Table \ref{tab:tab1} are calculated from the last 100 training epochs of each method. Comparing the metrics shown in Table 1, demonstrates that the proposed models significantly outperform the SNN method. Based on the data in Table \ref{tab:tab1} we can induce 67\% and 72\% reductions in the average cumulative delay and queue length for the policy gradient method and 68\% and 73\% reductions for value-function based method compared to the SNN. Furthermore, we can see that the proposed methods have received average rewards superior to the SNN. Considering these results, it is obvious that the policy gradient and value-function agents could learn the control policies better than the SNN approach.

\section{Conclusion}
\label{conclusion}
In this paper, we applied deep reinforcement learning algorithms with focusing on both policy and value-function based methods to traffic signal control problem in order to find optimal control policies of signalling, just by using raw visual input data of the traffic simulator snapshots. Our approaches have led to promising results and showed they could find more stable control policies compared to previous work of using deep reinforcement learning in traffic light optimization. In our work, we developed and tested the proposed methods in a small application, extending the work for more complex traffic simulations, for instance considering many intersections and multiple agents to control each intersection, using multi-agent learning techniques to handle coordination problem between agents would be a direction for future research.

\section*{Acknowledgments}
We would like to thank FotoNation company for letting us to work with their GPU cluster.

%
\bibliographystyle{abbrv}
\bibliography{aamas2017_sample.bib}  
%
\end{document}